\documentclass[letterpaper, 10 pt, conference]{ieeeconf}  %

\IEEEoverridecommandlockouts                              %

\usepackage[nocompress]{cite}
\usepackage{amsmath,amssymb,amsfonts}
\usepackage{algorithm}
\usepackage{algorithmic}
\usepackage{url}
\usepackage{booktabs,multirow}
\usepackage{bbm}
\floatstyle{ruled}
\restylefloat{algorithm}

\usepackage{graphicx}
\usepackage{textcomp}
\usepackage{xcolor}
\usepackage[hidelinks]{hyperref}
\title{\LARGE \bf
TimelyDAgger: Timing-Aware Expert Querying for VLA Policy Improvement
}

\author{Zhixuan Zhao$^{1,6,\dagger}$, Peiyan Li$^{1,6,\dagger}$, Enhao Zhang$^{2,6}$, Yueran Tao$^{1,6}$,\\
Hao Wang$^{3,6}$, Chenghao Yue$^{1,6}$, Lei Lv$^{4,6}$, Wentao Zhao$^{1,6}$, Jiahao Chen$^{5,6}$,\\
Xin Liu$^{1,6}$, Kangyao Huang$^{1,6}$, Yu Luo$^{1,6,*}$, Huaping Liu$^{1,6,*}$\thanks{$^{1}$Tsinghua University. $^{2}$Imperial College London. $^{3}$Dalian University of Technology. $^{4}$Tongji University. $^{5}$Peking University. $^{6}$SEEN\textperiodcentered E Robotics.}\thanks{$^{\dagger}$Zhixuan Zhao and Peiyan Li contributed equally and are co-first authors.}\thanks{$^{*}$Yu Luo and Huaping Liu are co-corresponding authors.}}

\begin{document}
\bstctlcite{compactrefs}

\maketitle
\thispagestyle{empty}

\pagestyle{empty}

\begin{abstract}
DAgger improves robot policies by aggregating expert supervision
from states visited during policy execution.
Robot-gated DAgger automates expert queries, allowing the robot
to decide when to request expert takeover.
While existing gates emphasize detecting the need for assistance,
takeover timing also shapes the content of these demonstrations
and their value for policy learning.
We propose TimelyDAgger, combining Bridge-PCA monitoring of internal vision-language-action (VLA) features with Feedback-guided Threshold Adaptation based on expert behavior to improve takeover timing. We introduce an evaluation framework linking failure detection, takeover timing, and policy improvement, including Target-Aligned Supervision Ratio (TASR) for assessing supervision quality without retraining. Experiments show that takeover timing affects policy learning, with TimelyDAgger achieving competitive failure detection and higher post-training success in most evaluated settings under matched expert-action budgets. Project website: \url{https://seen-e.github.io/TimelyDagger/}.
\end{abstract}

\begin{figure}[t]
    \centering
    \includegraphics[width=\columnwidth]{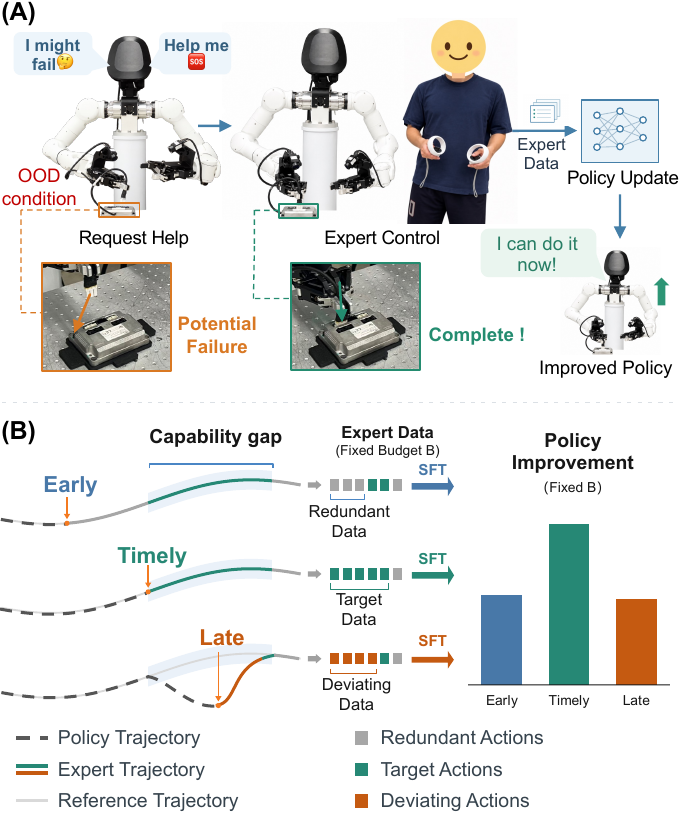}
\caption{
    \textbf{(A) Robot-gated DAgger.}
    The robot requests expert assistance and uses the resulting
    demonstrations to update its policy.
\textbf{(B) Takeover timing affects policy improvement.}
Early takeover can collect redundant behavior, while late
takeover may miss part of the behavior the policy needs to learn.
    Timely takeover concentrates supervision on the target behavior,
    supporting more effective policy learning.
}
    \label{fig:overview}
\end{figure}

\section{Introduction}

Recent advances in robot learning have produced policies capable of
performing a wide range of manipulation tasks.
Post-training with expert demonstrations plays an important role in
adapting these policies to specific tasks and deployment conditions,
but acquiring suitable demonstrations requires substantial expert
effort. An important challenge is therefore to collect additional
supervision that addresses the policy's remaining weaknesses while
keeping expert effort low.

DAgger provides a foundation for acquiring such supervision during
the robot's own execution: it queries an expert at states visited by
the policy and aggregates the resulting action labels with existing
demonstrations to improve the policy \cite{pmlr-v15-ross11a}.
Human-gated DAgger makes this process selective by allowing a human
supervisor to take control when correction is needed
\cite{kelly2019hgdagger}, \cite{mandlekar2020humanintheloopimitationlearningusing}.
Although this reduces unnecessary intervention, the supervisor must
still continuously monitor the robot.
Robot-gated DAgger automates the takeover decision using uncertainty
or failure signals, allowing the robot to request expert assistance
and collect corrective demonstrations without continuous human
monitoring \cite{pmlr-v164-hoque22a,lee2024diff,menda2019ensembledaggerbayesianapproachsafe}.

Existing robot gates primarily focus on detecting potential
policy failures and requesting expert help
\cite{pmlr-v164-hoque22a,lee2024diff}.
Reliable failure detection is an important component of
robot-gated DAgger, but detection metrics alone do not fully
characterize the learning value of the demonstrations collected
after takeover.
We ask how takeover timing affects the policy improvement
obtained from a given amount of expert supervision.
As illustrated in Fig.~\ref{fig:overview}, an unnecessarily
early takeover can still capture the behavior requiring
improvement, but may also include a long prefix of behavior
the policy already performs reliably.
Conversely, a late takeover may miss part of the behavior
the policy needs to learn, as the collected expert
demonstration begins only after execution has deviated
from the intended task trajectory.
Both interventions can complete the task successfully, yet
devote different proportions of expert supervision to the
policy's remaining capability gap.
Timely takeover can increase this proportion, potentially
supporting greater policy improvement from the same amount
of expert data.
This distinction motivates treating takeover timing as a data
acquisition decision and evaluating its effect on the data
efficiency of policy learning.

Building on this idea, we propose TimelyDAgger, a robot-gated
DAgger method that combines monitoring of internal policy features
with expert feedback to improve takeover timing.
For vision-language-action (VLA) policies, its Bridge-PCA component
monitors the features passed from the vision-language model to the
action generation module. A PCA subspace fitted to successful
demonstrations serves as a reference, and the robot requests help
when the feature reconstruction error exceeds a calibrated threshold.
Feedback-guided Threshold Adaptation (FTA) then uses feedback from
completed interventions to adjust the takeover threshold for
better timing in subsequent episodes. The expert controls the remainder
of the episode after takeover, and successful expert recordings are
combined with the original demonstrations to fine-tune the policy.

To evaluate how these takeover decisions affect learning, we introduce
a dedicated framework for robot-gated DAgger that connects failure
detection, takeover timing, and downstream policy improvement.
We further propose the Target-Aligned Supervision Ratio (TASR),
a metric for assessing the quality of collected expert supervision
without policy retraining.
Experiments covering 11 manipulation tasks show that takeover timing substantially affects policy
improvement, with TimelyDAgger achieving competitive failure
detection and higher downstream success than the alternatives evaluated 
in most reported settings.

Our contributions are threefold:
\begin{itemize}
\item \textbf{Formulation.}
We formulate \emph{expert takeover timing} as a
central problem in robot-gated DAgger. We demonstrate
that takeover timing substantially affects the content of collected
demonstrations and subsequent policy improvement.

\item \textbf{Method.}
We propose TimelyDAgger, a robot-gated DAgger method with two
components. Bridge-PCA monitors deviations in internal VLA features
to trigger expert takeover. Feedback-guided Threshold Adaptation (FTA)
uses expert behavior during completed interventions to adjust takeover
thresholds for better takeover timing.

\item \textbf{Evaluation.}
We introduce a dedicated evaluation framework for robot-gated DAgger,
covering failure detection, expert data collection, and downstream
policy learning. We further propose TASR, a simple but effective metric for assessing the quality of collected expert supervision
without policy retraining.

\end{itemize}
\section{Related Work}

\subsection{Active Imitation Learning and Robot-Gated DAgger}

Imitation learning trains policies from expert demonstrations but remains
vulnerable to distribution shifts during deployment. Active imitation learning
addresses this limitation by selectively acquiring expert supervision on
states visited by the learned policy \cite{Chernova_2009,JMLR:v15:judah14a}. DAgger is a foundational approach that
iteratively queries expert actions on learner-induced states and aggregates
the resulting supervision into the training set
\cite{pmlr-v15-ross11a}. Human-gated DAgger allows a supervisor to intervene
only when correction is needed, but still requires continuous human monitoring
and timely control \cite{kelly2019hgdagger}. Robot-gated DAgger instead
automates the intervention decision using a learned or uncertainty-based gate.

Existing robot-gated DAgger methods construct gates from safety classifiers
\cite{zhang2016queryefficientimitationlearningendtoend}, disagreement among policy ensembles
\cite{menda2019ensembledaggerbayesianapproachsafe}, or combined novelty and risk estimates
\cite{pmlr-v164-hoque22a}. 
More recent approaches exploit signals from generative action policies:
Diff-DAgger uses diffusion training loss to trigger expert intervention
\cite{lee2024diff}, while \emph{Uncertainty Comes for Free} derives an
uncertainty signal from the diffusion denoising process \cite{he2025uncertaintycomesfreehumanintheloop}.
AutoIntervene combines visual support with action consistency to calibrate
control transfer for action-chunking policies
\cite{tang2026autointervene}. These methods primarily focus on reliably determining
whether expert intervention is needed, while rarely examining takeover timing
as a separate factor. We instead treat timing as a data-acquisition decision that determines
where corrective demonstrations begin, and examine how these
demonstrations improve policy performance after fine-tuning.

\subsection{Runtime Monitoring as Robot-Gating Signals}

Robot-gated DAgger requires a monitoring signal to decide when to request
expert assistance. Methods from the broader literature on runtime failure
and OOD detection provide several sources of such signals.
Observation-based approaches detect departures from nominal inputs using
feature distances or learned density models~\cite{NEURIPS2018_abdeb6f5,
pmlr-v162-sun22d,XuC2-RSS-25,Zhou_2026_CVPR}.
Internal-feature approaches learn failure predictors from representations
computed within the VLA~\cite{NEURIPS2025_392d0d05,
park2026hideandseektrajectoriesdiscoveringfailure}.
Action-based approaches estimate uncertainty or inconsistency through
policy ensembles, repeated action sampling, generative-model losses,
or overlapping action chunks~\cite{
menda2019ensembledaggerbayesianapproachsafe,pmlr-v270-agia25a,
NEURIPS2025_0b7cb3b8,lee2024diff,
he2025uncertaintycomesfreehumanintheloop,
zheng2026rewindilonlinefailuredetection}.
Semantic approaches use VLMs to assess task failure or progress,
either directly or together with visual anomaly detection~\cite{lin2026failsafe,ma2026cyclevla}.

However, the computational and supervision costs of these
approaches can limit their suitability for robot-gated DAgger.
External reasoning models~\cite{ICLR2025_70a06501} and
repeated action sampling~\cite{pmlr-v270-agia25a} add runtime
computation, while learned failure detectors require additional
training~\cite{NEURIPS2025_0b7cb3b8} and, in some cases,
labeled failure rollouts~\cite{NEURIPS2025_392d0d05}.
When obtaining these labels requires expert input, the added
supervision cost can partially offset the expert-effort savings
sought by robot-gated DAgger.
Our approach instead reuses existing VLA features, training
demonstrations, and feedback from interventions collected for
policy learning, requiring neither additional predictor training
nor dedicated data collection for the gate.

\begin{figure*}[t]
    \centering
    \includegraphics[width=\textwidth]{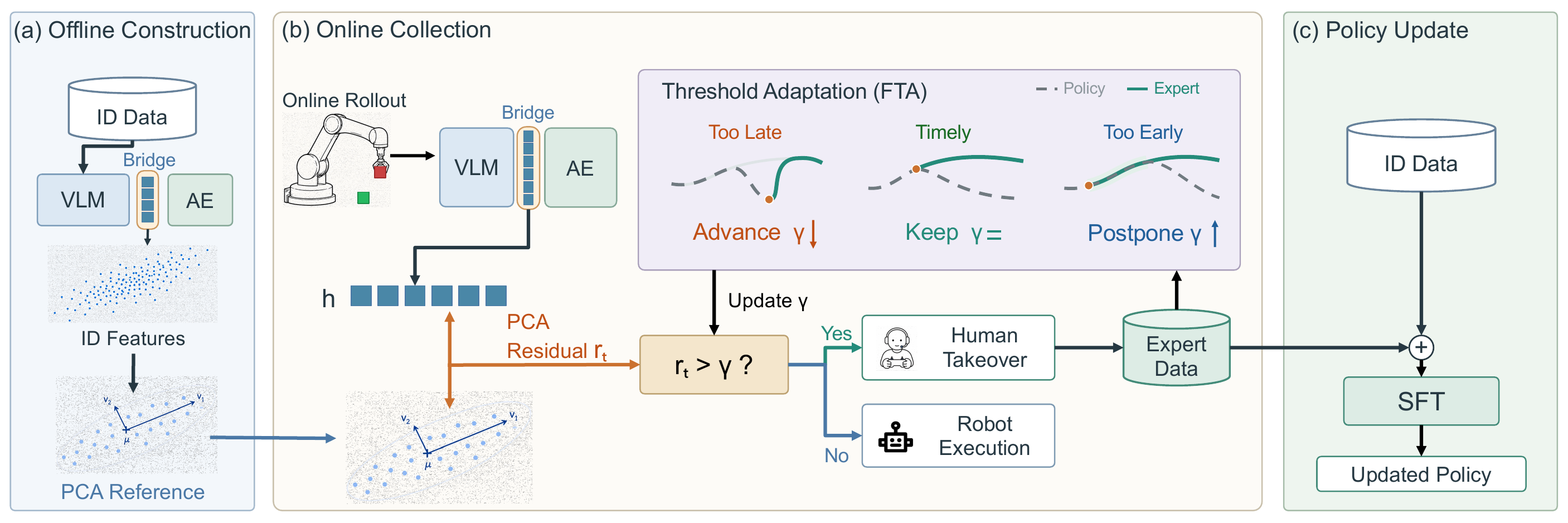}
    \caption{
        \textbf{Overview of TimelyDAgger.}
        \textbf{(a) Offline construction:}
        Bridge-PCA fits a principal subspace to ID bridge features
        and calibrates an initial takeover threshold.
        \textbf{(b) Online collection:}
        Bridge-PCA monitors policy execution and requests expert
        takeover, while FTA uses intervention feedback to adjust
        the threshold for better timing in subsequent episodes.
        \textbf{(c) Policy update:}
        The collected expert suffixes are combined with the original
        ID demonstrations to fine-tune the policy.
    }
    \label{fig:method_double}
\end{figure*}

\subsection{Takeover Timing and Supervision Quality}

Prior work has recognized that intervention timing affects the value of expert
supervision. Intervention Weighted Regression observes that human interventions
concentrate around task bottlenecks and prioritizes these segments during
training \cite{mandlekar2020humanintheloopimitationlearningusing}. LazyDAgger and ThriftyDAgger regulate
intervention timing and duration to balance task performance with supervisor
burden \cite{hoque2021lazydagger,pmlr-v164-hoque22a}. AutoIntervene further
argues that unnecessarily prolonged interventions introduce already-supported
behavior and dilute corrective updates \cite{tang2026autointervene}.
However, intervention accuracy and supervisor burden alone do not reveal
whether takeover occurs at a useful point for collecting training data.
They provide limited insight into whether the collected
demonstrations adequately cover the behavior needed to
address the policy's capability gap. We explicitly formulate expert takeover timing
as a data acquisition problem and design TimelyDAgger to refine takeover
decisions using feedback from completed expert interventions. We also
introduce a dedicated evaluation framework and TASR to examine how
takeover timing shapes expert supervision and subsequent policy
improvement.

\section{Problem Formulation}
\label{sec:problem_formulation}

Let $\pi_\theta$ be a robot policy trained on demonstrations
$\mathcal D_{ID}$, and let $\pi^\ast$ be an expert.
During a rollout $\tau$, a robot gate $g$ decides whether to request
expert assistance. We write $Q_g(\tau)=1$ when assistance is requested
and $Q_g(\tau)=0$ when the rollout remains autonomous.
For an assisted rollout, $T_g(\tau)$ denotes the time at which
control transfers from the policy to the expert:
\begin{equation}
    a_t \sim
    \begin{cases}
        \pi_\theta(\cdot\mid o_t), & t<T_g(\tau),\\
        \pi^\ast(\cdot\mid o_t), & t\geq T_g(\tau).
    \end{cases}
\end{equation}
The expert completes the episode, and its observation-action pairs
from $T_g(\tau)$ onward form the corrective demonstration.
Thus, $Q_g$ determines which executions contribute expert data,
while $T_g$ determines where each demonstration begins.

Let $\mathcal D_E(Q,T;B)$ denote the expert data collected under
selection rule $Q$ and timing rule $T$, with $B$ expert actions
retained for training. We define their learning utility as
\begin{equation}
\begin{aligned}
    \pi_{\theta'}
    &= \operatorname{Update}\left(
        \pi_\theta,\,
        \mathcal D_{\mathrm{ID}}\cup\mathcal D_E(Q,T;B)
    \right),\\
    U_B(Q,T)
    &= J(\pi_{\theta'}),
\end{aligned}
\end{equation}
where $J$ measures the updated policy's performance.

With the selection rule $Q$, base policy, expert action budget,
and training procedure fixed, we prefer timing rule $T_1$ to $T_2$
when its demonstrations produce higher success after fine-tuning:
\begin{equation}
    T_1\succ_{B,Q}T_2
    \quad\Longleftrightarrow\quad
    U_B(Q,T_1)>U_B(Q,T_2).
\end{equation}
For a complete robot gate, we evaluate the combined effect of
execution selection and takeover timing through $U_B(Q_g,T_g)$.

\begin{figure}[t]
    \centering
    \includegraphics[width=\columnwidth]{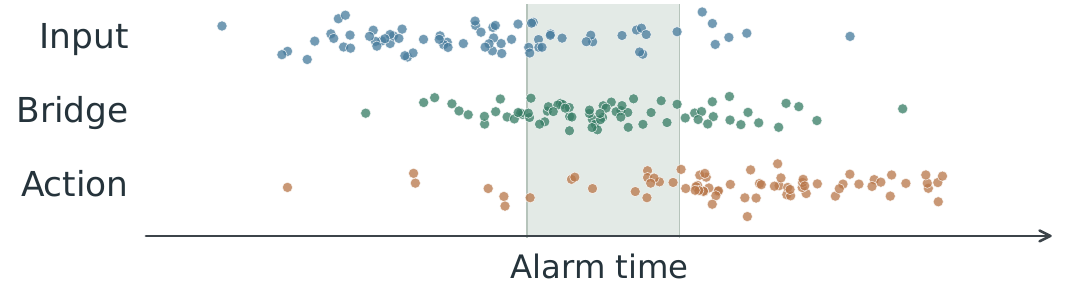}
    \caption{
        Comparison of alarm timing across monitoring locations.
    A larger proportion of bridge alarms falls within the shaded
    preferred takeover window than is observed for input and action alarm, indicating closer alignment with
    the desired intervention timing.
    }
    \label{fig:bridge_timing}
\end{figure}

\section{Method}
\label{sec:method}

TimelyDAgger combines Bridge-PCA and Feedback-guided Threshold
Adaptation (FTA) in three stages (Fig.~\ref{fig:method_double}):
\textbf{(1) Offline gate construction:} fit a PCA subspace to
ID bridge features and calibrate a takeover threshold;
\textbf{(2) Online expert data collection:} trigger takeover
when the residual exceeds the threshold and adapt the threshold
using intervention feedback for subsequent episodes;
\textbf{(3) Policy update:} fine-tune the policy on retained
expert suffixes and original ID demonstrations.
Algorithm~\ref{alg:introspective_dagger} summarizes the complete procedure.

\subsection{Offline Gate Construction}
\label{sec:bridge_pca}

\paragraph{ID Reference Features}
We monitor the bridge representation passed from the VLM to the
action generation module, where visual observations have been
processed in the context of the task instruction. This makes the
bridge a useful candidate for detecting changes relevant to the
required behavior: input novelty can appear before assistance is
needed, while action inconsistency can emerge after an incorrect
action has been produced. Fig.~\ref{fig:bridge_timing} illustrates
this timing motivation.

Let $z_t^{\mathrm B}=f_\theta^{\mathrm B}(o_t)\in\mathbb R^d$
denote the bridge feature at policy-query time $t$.
Using the frozen policy, we extract these features from observations
in $\mathcal D_{\mathrm{ID}}$ to construct the reference set
$\mathcal Z_{\mathrm{ID}}^{\mathrm B}=\{z_i^{\mathrm B}\}_{i=1}^{N}$,
where each sample corresponds to one observation.

\paragraph{Principal-Subspace Residual}

\begin{figure}[t]
    \centering
    \includegraphics[width=\columnwidth]{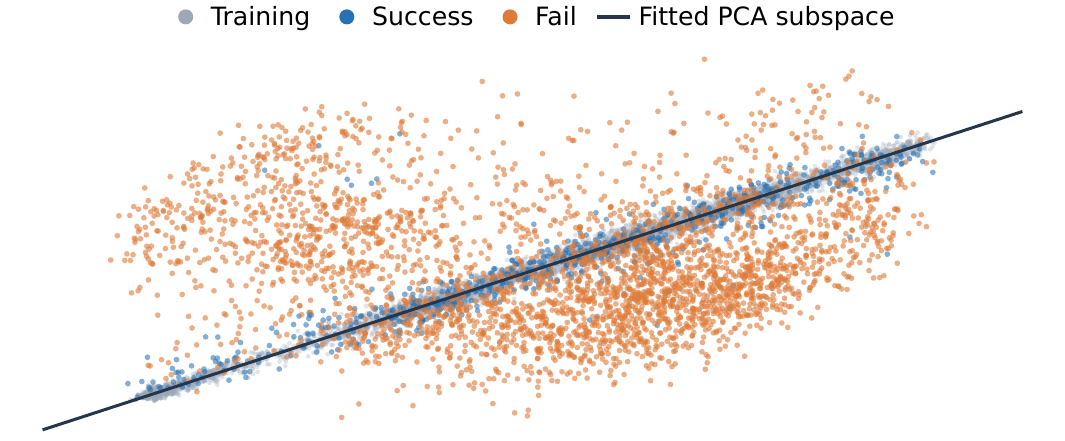}
    \caption{
        Bridge-feature geometry. Gray points denote ID reference features;
        blue and orange points denote features from successful and failed
        policy executions. The dark line illustrates the fitted PCA
        subspace in the visualization. The separation relative to this
        reference motivates the residual used by Bridge-PCA.
    }
    \label{fig:bridge_pca_geometry}
\end{figure}

Fig.~\ref{fig:bridge_pca_geometry} shows that features from successful
executions concentrate near the ID reference structure, whereas
features from failed executions exhibit larger deviations.
Motivated by this geometry, we fit PCA to capture the dominant
variation of the reference features and measure departures from
the resulting subspace. The empirical mean and covariance are
\begin{equation}
\begin{aligned}
    \mu_{\mathrm B}
    &= \frac{1}{N}\sum_{i=1}^{N}z_i^{\mathrm B},\\
    \Sigma_{\mathrm B}
    &= \frac{1}{N}\sum_{i=1}^{N}
    (z_i^{\mathrm B}-\mu_{\mathrm B})
    (z_i^{\mathrm B}-\mu_{\mathrm B})^\top.
\end{aligned}
\end{equation}
Let $V_{\mathrm B,r}\in\mathbb R^{d\times r}$ contain the orthonormal
eigenvectors associated with the $r$ largest eigenvalues of
$\Sigma_{\mathrm B}$, and define
$P_{\mathrm B}=V_{\mathrm B,r}V_{\mathrm B,r}^{\top}$.
The affine subspace
$\mu_{\mathrm B}+\operatorname{span}(V_{\mathrm B,r})$ serves as the ID
reference, with rank $r<d$ fixed before evaluation. We define the monitoring
score as
\begin{equation}
    s_{\mathrm{BPCA}}(o_t)
    = \left\|(I-P_{\mathrm B})
    (z_t^{\mathrm B}-\mu_{\mathrm B})\right\|_2.
    \label{eq:bridge_pca_score}
\end{equation}
The residual measures the component of the current feature that
cannot be reconstructed from the ID principal directions.
A larger residual therefore indicates a greater departure from
the reference subspace and provides the monitoring score used
for threshold calibration.

\paragraph{Threshold Calibration}
To turn the residual into an intervention decision, we calibrate a threshold
on successful ID data $\mathcal D_{\mathrm{cal}}$. For each calibration rollout, we
compute its maximum residual
$m(\tau)=\max_t s_{\mathrm{BPCA}}(o_t)$ and set
\begin{equation}
    \gamma_{\mathrm B}
    = \operatorname{Quantile}_{1-\alpha}
    \left(\{m(\tau):\tau\in\mathcal D_{\mathrm{cal}}\}\right),
    \label{eq:bridge_pca_threshold}
\end{equation}
where $\alpha$ specifies the upper-tail calibration level. The resulting $\gamma_{\mathrm B}$ provides
the baseline threshold for online collection.

\subsection{Online Expert Data Collection}

Bridge-PCA provides the initial takeover threshold.
During data collection, Feedback-guided Threshold Adaptation (FTA)
uses completed expert interventions to adjust this threshold for
subsequent episodes to optimize takeover timing of the gate.

\subsubsection{Expert Takeover}

We initialize the threshold to $\gamma_1=\gamma_{\mathrm B}$.
At each policy query, the robot computes $s_{\mathrm{BPCA}}(o_t)$.
If the score exceeds the current threshold, the expert takes control
and completes the episode. Otherwise, the robot executes the next
$\Delta$ policy actions before checking again.
Successful expert suffixes are retained in
$\mathcal D_{\mathrm{corr}}$ for policy updating.

\begin{algorithm}[t]
\caption{TimelyDAgger}
\label{alg:introspective_dagger}
\begin{algorithmic}[1]
\REQUIRE Base policy $\pi_\theta$, expert $\pi^\ast$,
ID demonstrations $\mathcal D_{\mathrm{ID}}$,
ID calibration data $\mathcal D_{\mathrm{cal}}$;
execution block length $\Delta$, FTA step size $\beta$

\STATE Extract bridge features from $\mathcal D_{\mathrm{ID}}$ using $\pi_\theta$
\STATE Fit Bridge-PCA to obtain $s_{\mathrm{BPCA}}$
\STATE Calibrate $\gamma$ on $\mathcal D_{\mathrm{cal}}$
\STATE Initialize expert dataset $\mathcal D_{\mathrm{corr}}\gets\emptyset$

\FOR{each collection episode}
    \STATE Reset environment
    \WHILE{episode not terminated}
        \STATE Observe $o$
        \IF{$s_{\mathrm{BPCA}}(o)>\gamma$}
            \STATE Execute $\pi^\ast$ to episode end; record suffix $\xi$
            \STATE Compute timing cue $y$ using Eq.~\eqref{eq:timing_cue}
            \STATE $\gamma\gets\gamma\exp(-\beta y)$
            \STATE $\mathcal D_{\mathrm{corr}}\gets
            \mathcal D_{\mathrm{corr}}\cup\{\xi\}$
            \STATE \textbf{break}
        \ELSE
            \STATE Execute the next $\Delta$ actions from $\pi_\theta$
        \ENDIF
    \ENDWHILE
\ENDFOR

\STATE Fine-tune $\pi_\theta$ on
$\mathcal D_{\mathrm{ID}}$ and $\mathcal D_{\mathrm{corr}}$
\RETURN $\pi_\theta$
\end{algorithmic}
\end{algorithm}

\subsubsection{Feedback-guided Threshold Adaptation (FTA)}

\paragraph{Timing Feedback}
FTA uses the first $\Delta$ expert actions after takeover to assess
whether assistance was requested too early or too late.
During this interval, the frozen policy also predicts actions from
the same observations, while only the expert actions are executed.
We extract two feedback signals.

\textbf{(1) Corrective reversal.}
We compare the robot's motion immediately
before takeover with the expert's initial motion afterward.
For example, if the robot moves forward and the expert immediately
moves it back, the intervention begins by undoing the preceding motion.
We set $c_i=1$ when this reversal exceeds the correction tolerance,
and $c_i=0$ otherwise. A substantial reversal suggests that execution had deviated
before takeover and provides a heuristic cue for requesting
assistance earlier.

\textbf{(2) Policy-expert agreement.}
At each observation during the initial expert block, we compare
the policy's predicted action with the action executed by the expert.
We set $u_i=1$ if their differences remain within the agreement
tolerance throughout the block, and $u_i=0$ otherwise.
Agreement indicates that the policy would have chosen similar actions
at these observations, suggesting that takeover could be postponed
when no substantial correction is observed.

For completed intervention $i$, we combine these signals into
\begin{equation}
    y_i =
    \begin{cases}
        +1, & c_i=1
              \quad \text{(advance)},\\
        -1, & c_i=0 \ \text{and}\ u_i=1
              \quad \text{(postpone)},\\
        0,  & \text{otherwise}
              \quad \text{(keep)}.
    \end{cases}
    \label{eq:timing_cue}
\end{equation}
Correction takes priority: agreement after takeover does not cancel
evidence that the expert first had to undo the robot's motion.
The resulting cue adjusts the threshold for subsequent episodes.

\paragraph{Threshold Update}
After the intervention ends, we update the threshold as
\begin{equation}
    \gamma_{i+1}
    = \gamma_i \exp(-\beta y_i),
    \qquad \beta>0,
    \label{eq:fta_threshold}
\end{equation}
where $\beta$ controls the adjustment size.
An advance cue lowers the threshold, making assistance easier
to trigger; a postpone cue raises it, allowing more autonomous
execution. The updated threshold applies to all states in the
current task until another completed intervention supplies feedback.
Episodes without intervention leave it unchanged. A real-world example in Fig.~\ref{fig:real_world_tasks}
illustrates how FTA lowers the threshold in response to expert
correction, enabling earlier takeover in a subsequent rollout.

\subsection{Policy Update}
\label{sec:policy_update}

After data collection, we fine-tune the base policy using the retained
expert demonstrations $\mathcal D_{\mathrm{corr}}$ together with the
original ID demonstrations $\mathcal D_{\mathrm{ID}}$.
Each mini-batch samples from both datasets at a fixed ratio. The updated policy can then serve as the base policy for the next
round, repeating gate construction, expert data collection,
and policy updating.

\begin{figure}[t]
    \centering
    \includegraphics[width=\columnwidth]{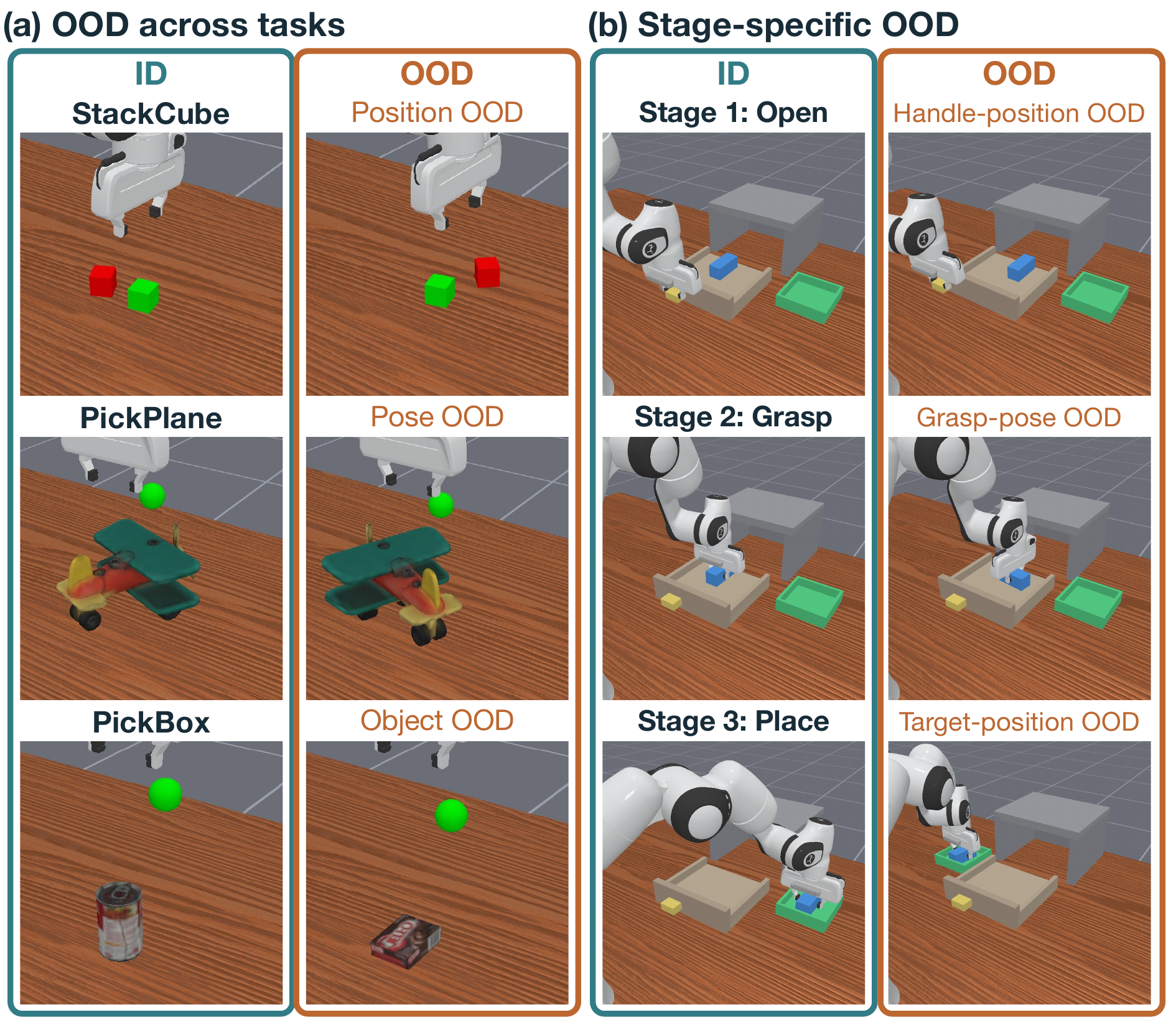}
    \caption{
        ID-to-OOD task design.
        (a) Representative position, pose, and object variations across tasks.
        (b) Variations affecting different stages of OpenDrawer:
        handle position for opening, object pose for grasping, and
        target position for placement. Each pair contrasts the ID
        configuration with its corresponding OOD variation.
    }
    \label{fig:task_design}
\end{figure}

\begin{table*}[t]
\centering
\caption{Simulation results across tasks and VLA backbones.}
\label{tab:simulation_main_results}
\footnotesize
\setlength{\tabcolsep}{2pt}
\renewcommand{\arraystretch}{1.16}
\begin{tabular*}{\textwidth}{
    @{\hspace{2pt}\extracolsep{\fill}}
    ll
    *{6}{c}
    @{\hspace{8pt}}
    *{4}{c}
    @{\hspace{8pt}}
    *{4}{c}
    @{\hspace{2pt}}
}
\toprule
\multirow{2}{*}{VLA} & \multirow{2}{*}{Task} & \multicolumn{6}{c}{AUPRC (\%) $\uparrow$} & \multicolumn{4}{c}{TASR (\%) $\uparrow$} & \multicolumn{4}{c}{Success Rate (\%) $\uparrow$} \\
\cmidrule(lr){3-8}\cmidrule(lr){9-12}\cmidrule(lr){13-16}
 & & FID & CRS & Diff & ACC & STAC & Ours & BC & HG & Diff & Ours & BC & HG & Diff & Ours \\
\midrule
\multirow{11}{*}{$\pi_{0.5}$} & SC--Green & 96.74 & 94.67 & 80.88 & 72.11 & 63.82 & \textbf{99.72} & 49.24 & 24.22 & 22.20 & \textbf{87.82} & 49 & 29 & 34 & \textbf{81} \\
 & SC--Red & 98.36 & \textbf{99.57} & 82.94 & 71.13 & 65.87 & 99.30 & 35.01 & 35.36 & 31.66 & \textbf{46.25} & 48 & 39 & 55 & \textbf{76} \\
 & OD--Handle & 96.82 & 98.70 & 83.32 & 54.31 & 60.69 & \textbf{98.80} & 18.45 & 19.85 & 20.59 & \textbf{24.71} & 46 & 41 & 55 & \textbf{68} \\
 & OD--Place & 96.66 & \textbf{98.30} & 87.79 & 68.77 & 63.94 & 97.73 & 17.46 & 18.89 & 26.41 & \textbf{28.50} & 43 & 37 & 52 & \textbf{64} \\
 & SP--Green & 93.25 & 94.33 & 89.94 & 64.25 & 73.67 &\textbf{97.35} & 28.24 & 25.85 & 26.52 & \textbf{33.49} & 50 & 47 & 46 & \textbf{53} \\
 & SP--Red & 84.64 & 91.46 & 80.20 & 76.03 & 55.62 & \textbf{92.70} & 27.54 & 24.31 & 26.47 & \textbf{45.66} & 60 & 57 & 45 & \textbf{79} \\
 & SP--Blue & 98.87 & 96.01 & 90.22 & 79.63 & 69.31 & \textbf{98.94} & 24.13 & 17.26 & 20.91 & \textbf{30.88} & 75 & 67 & 54 & \textbf{80} \\
\cmidrule(l){2-16}
 & PickPlane & 89.77 & 88.51 & 86.99 & 73.11 & 78.54 & \textbf{96.35} & 26.52 & 59.87 & 68.52 & \textbf{72.24} & 0 & 63 & \textbf{84} & 81 \\
 & OD--Pose & 90.89 & 91.89 & 86.70 & 78.29 & 65.98 & \textbf{98.78} & 25.04 & 12.78 & 29.36 & \textbf{37.28} & 42 & 35 & 51 & \textbf{66} \\
\cmidrule(l){2-16}
 & YCB--Object & 97.73 & 99.27 & 97.96 & 88.77 & 65.06 & \textbf{99.69} & 49.27 & 72.08 & 63.89 & \textbf{91.84} & 48 & 49 & 42 & \textbf{52} \\
 & Eggplant & 74.48 & 78.63 & 71.47 & 62.35 & 59.72 & \textbf{86.90} & 49.36 & 43.20 & \textbf{78.35} & 76.20 & 47 & 40 & 55 & \textbf{69} \\
\midrule
\multirow{2}{*}{OpenVLA} & PickPlane & 88.00 & 88.42 & -- & -- & -- & \textbf{90.73} & 50.56 & 36.20 & 38.01 & \textbf{58.25} & 30 & 26 & 14 & \textbf{50} \\
 & SC-Green & 96.56 & 93.32 & -- & -- & -- & \textbf{98.36} & 45.39 & 35.06 & 41.69 & \textbf{52.88} & 71 & 63 & 64 & \textbf{82} \\
\midrule
\multirow{2}{*}{X-VLA} & PickPlane & 53.42 & 55.30 & 55.81 & 29.41 & 27.21 & \textbf{61.30} & 56.21 & 30.58 & 32.89 & \textbf{59.82} & 72 & 35 & 58 & \textbf{84} \\
 & SC-Green & 97.75 & 99.66 & 95.03 & 67.37 & 55.25 & \textbf{100.00} & \textbf{54.68} & 53.69 & 46.73 & 53.90 & 58 & \textbf{76} & 64 & 67 \\
\bottomrule
\end{tabular*}
\par\vspace{4pt}
\begin{minipage}{\textwidth}
\scriptsize
FID: FIDeL; CRS: CRSAIL; Diff: Diff-DAgger; BC: Offline BC; HG: Human-Gated DAgger. SC, OD, and SP denote StackCube, OpenDrawer, and StackPyramid.
\end{minipage}
\end{table*}

\section{Experiments}
\label{sec:experiments}

Our experiments test three hypotheses:

\textbf{H1: Takeover Timing.}
Different takeover times change the composition of expert demonstrations
and the resulting policy improvement.

\textbf{H2: TimelyDAgger Effectiveness.}
TimelyDAgger reliably identifies executions requiring expert assistance
and collects demonstrations that lead to greater policy improvement
than the evaluated alternatives.

\textbf{H3: Supervision Quality.}
TASR serves as a simple but effective metric for assessing the
quality of expert supervision without policy retraining,
with higher TASR indicating greater downstream policy improvement.

The remainder of this section contains three parts.
\textbf{(1) Simulation Evaluation Framework} introduces the tasks, evaluation metrics,
baselines, and experimental protocol.
\textbf{(2) Simulation Results} presents the experimental findings.
\textbf{(3) Real-World Validation} evaluates the method on a physical robot.

\subsection{Simulation Evaluation Framework}

\subsubsection{ID-to-OOD Task Design}
\label{sec:task_design}

We construct 11 manipulation task variants in ManiSkill \cite{taomaniskill3}. Each base
policy is trained on a restricted ID distribution and deployed under
OOD conditions that change \textbf{(1) object position}, \textbf{(2) object pose}, or
\textbf{(3) object identity}, while
preserving the task objective and robot interface.
Fig.~\ref{fig:task_design}(a) shows representative variations.
We also vary the stage at which these changes affect the required
behavior, so that the need for expert assistance arises at different
times across tasks. For example, changes to the handle position,
grasp pose, and placement target in OpenDrawer affect opening,
grasping, and placement, respectively
(Fig.~\ref{fig:task_design}(b)). This temporal variation allows us
to evaluate whether a gate selects appropriate takeover times when
assistance is needed early, midway through, or late in an execution.
Table~\ref{tab:simulation_main_results} lists all tasks.

\begin{figure}[t]
    \centering
    \includegraphics[width=\columnwidth]{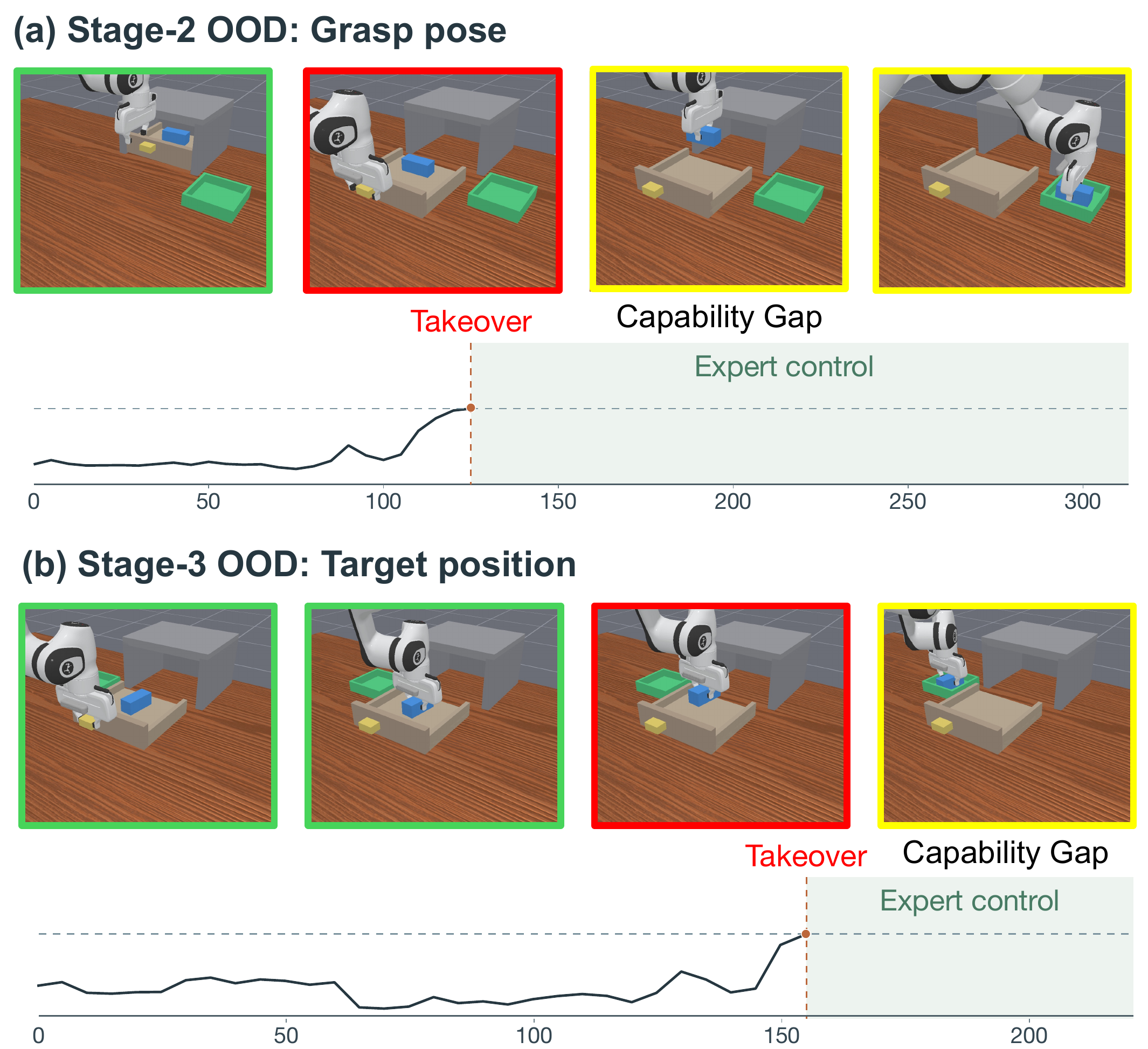}
    \caption{Qualitative illustrations and recorded monitoring scores
    for OpenDrawer under (a) grasp-pose OOD and (b) target-position OOD. These examples illustrate the goal of requesting expert assistance
before the task stage associated with the capability gap,
so that the collected demonstrations better support learning
the target behavior.}
    \label{fig:simulation_qualitative}
\end{figure}

\begin{figure*}[t]
    \centering
    \includegraphics[width=\textwidth]{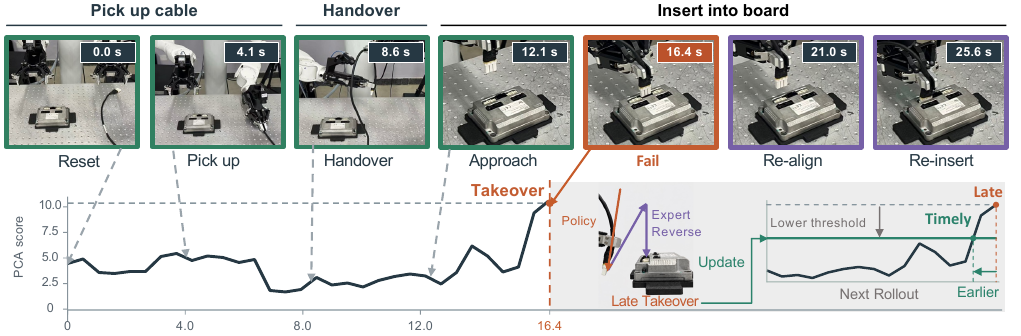}
\caption{
    \textbf{Real-world task and an illustrative FTA case.}
    \textbf{Top:} The task comprises cable pickup, handover,
    and precise connector insertion into the target board.
    \textbf{Bottom:} Expert correction following a late takeover
    provides feedback for FTA to lower the takeover threshold,
    enabling an earlier request for assistance in the next rollout.
}
    \label{fig:real_world_tasks}
\end{figure*}

\subsubsection{Evaluation Metrics}
\label{sec:evaluation_metrics}

\noindent\textbf{(1) Failure detection.}
We evaluate monitoring reliability using AUPRC
\cite{Davis_2006} on shared autonomous rollouts.

\noindent\textbf{(2) Policy improvement.}
We measure downstream learning performance by the updated policy's
success rate on OOD conditions.

\noindent\textbf{(3) Supervision quality.}
We propose the Target-Aligned Supervision Ratio (TASR), a metric
for evaluating expert takeover timing and the quality of collected
demonstrations without policy retraining.

\emph{Reference construction.}
For each condition $c$, the controlled ID-to-OOD design identifies
the task operation affected by the distribution shift.
We use successful OOD expert trajectories under the same condition
as references and manually annotate the interval corresponding
to this operation.
Let $\mathcal I_c$ denote this target interval in the reference
trajectory used for comparison.

\emph{Alignment and scoring.}
Each collected expert suffix $\xi_i$ is aligned with a reference
trajectory in execution order, allowing different execution speeds.
We set $\chi_i(t)=1$ when the matched reference point belongs to
$\mathcal I_c$ and the recorded state falls within a calibrated
tolerance of the reference state; otherwise, $\chi_i(t)=0$.
For the retained successful expert suffixes
$\mathcal D_g^{\mathrm{succ}}$, TASR is
\begin{equation}
Q_{\mathrm{TASR}}(g)
=
\frac{
\sum_{\xi_i\in\mathcal D_g^{\mathrm{succ}}}
\sum_{t=1}^{|\xi_i|}\chi_i(t)
}{
\sum_{\xi_i\in\mathcal D_g^{\mathrm{succ}}}|\xi_i|
},
\end{equation}
where $|\xi_i|$ is the number of retained expert actions
in suffix $\xi_i$, and $t$ indexes actions within the suffix.
The denominator counts all retained expert actions, while the
numerator excludes redundant actions outside the target interval
and deviations beyond the reference tolerance.
TASR thus characterizes the collected supervision without
policy retraining.

\subsubsection{Baselines}

We compare the Bridge-PCA monitor with FIDeL \cite{rolland2026failure}, CRSAIL \cite{firouzkouhi2025sampleefficientexpertquerycontrol}, Diff-DAgger~\cite{lee2024diff},
ACC \cite{seligmann2026vlafailefficienttaskfailure}, and STAC \cite{pmlr-v270-agia25a} using AUPRC. For expert data collection and policy
learning, we compare TimelyDAgger with Offline BC, Human-Gated DAgger,
and Diff-DAgger \cite{lee2024diff}, reporting TASR and post-training OOD success.
Offline BC records complete expert demonstrations, while the
interactive methods collect expert suffixes after takeover.

\subsubsection{Evaluation Protocol}

We evaluate $\pi_{0.5}$ \cite{intelligence2025pi05visionlanguageactionmodelopenworld}, OpenVLA \cite{kim24openvla}, and X-VLA \cite{zheng2025x}, with all methods starting
from the same ID-trained checkpoint within each task and backbone.
Monitors are calibrated on separate ID rollouts and evaluated on
shared autonomous ID/OOD trajectories, treating failure as the
positive class for AUPRC. We then collect expert data under each
acquisition rule. A task-specific oracle replaces human control
and is assumed to complete the task from the evaluated takeover
states; a predefined failure-triggered rule approximates human
takeover for Human-Gated DAgger. Successful expert recordings are
retained and evaluated using TASR. For policy updating, we match
the retained expert-action budget, the sampling ratio with original
ID data, and the training settings. Updated policies are tested
on OOD configurations without assistance.

\begin{figure}[t]
    \centering
    \includegraphics[width=\columnwidth]
    {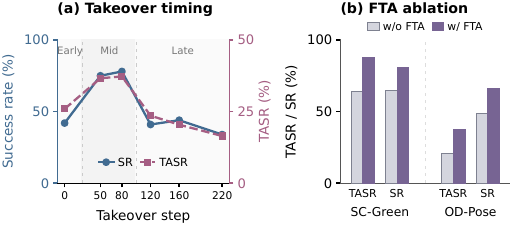}
    \caption{
        \textbf{(a) Takeover Timing Study.}
        Post-training success rate and TASR
        across six fixed takeover steps on
        OpenDrawer (Grasp-OOD) with matched retained expert-action budgets and training settings.
        \textbf{(b) FTA Ablation.}
        TASR and post-training success with and without FTA
        on StackCube--Green and OpenDrawer--Pose. 
    }
    \label{fig:timing_fta}
\end{figure}

\begin{figure}[t]
    \centering
    \includegraphics[width=\columnwidth]
    {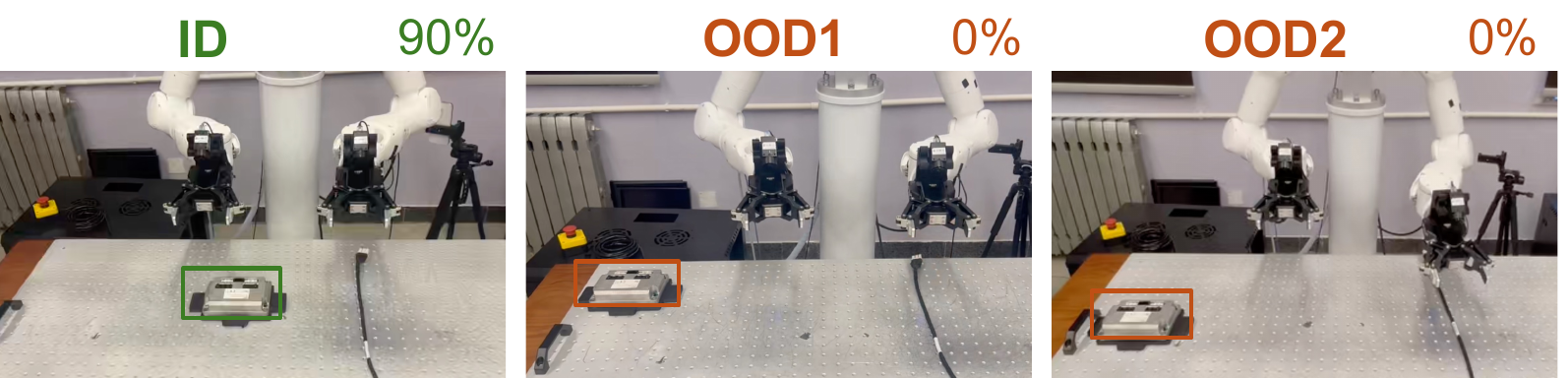}
    \caption{Real-world ID and OOD configurations for cable
    insertion. The target board is fixed at its training
    position in ID and moved to two different positions
    in OOD1 and OOD2 for data collection and evaluation.
    The percentage at the upper right of each panel
    indicates the base policy's success rate.}
    \label{fig:real_world_setup}
\end{figure}

\begin{figure}[t]
    \centering
    \includegraphics[width=\columnwidth]{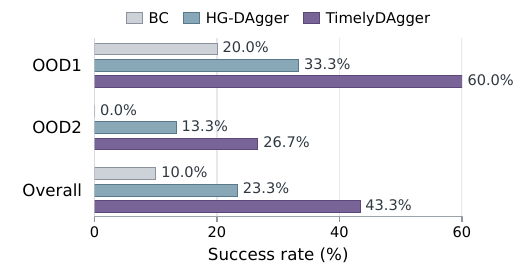}
    \caption{Real-world success rates under two OOD conditions.
    Overall combines both conditions.}
    \label{fig:real_world_success}
\end{figure}

\subsection{Simulation Results}
\label{sec:simulation_results}

We first report the main results obtained using the proposed
evaluation framework.
We then present two additional experiments: a controlled study
of takeover timing and an ablation of Feedback-guided Threshold
Adaptation.

\subsubsection{Main Evaluation of Robot-Gated DAgger}

Across the 15 task--backbone settings in
Table~\ref{tab:simulation_main_results}, TimelyDAgger achieves
the highest reported AUPRC, TASR, and post-training success
in 13 settings for each metric.
These results support H2, demonstrating competitive detection
performance and effective expert data collection for policy learning.
Spearman correlations between TASR and success, computed across
the four methods within each setting, average
$\bar{\rho}_s=0.72$ and are positive in 14 of 15 settings,
supporting H3.
Qualitative examples of takeover behavior are shown in
Fig.~\ref{fig:simulation_qualitative}.

\subsubsection{Controlled Study of Takeover Timing}

We collect expert suffixes at six fixed takeover steps on
OpenDrawer (Grasp-OOD).
Copies of the same base policy are fine-tuned on these datasets
using matched expert action budgets and training settings.
Fig.~\ref{fig:timing_fta}(a) shows that post-training success
increases from $42\%$ with immediate takeover to $78\%$ at
step $80$.
These results support H1: takeover timing affects the
downstream success achieved with the same amount of expert
training data. The accompanying TASR trend is consistent
with more efficient allocation of supervision to the
behavior requiring improvement.

\subsubsection{Ablation of Feedback-guided Threshold Adaptation}

We compare TimelyDAgger with and without FTA on StackCube--Green
and OpenDrawer--Pose.
Without FTA, the calibrated takeover threshold remains fixed
during collection; the expert action budget and subsequent
training settings are identical between variants.
Fig.~\ref{fig:timing_fta}(b) shows higher TASR and downstream
success with FTA on both tasks, supporting the use of expert
feedback to improve the learning value of collected demonstrations.

\subsection{Real-World Validation}
\label{sec:real_world}

We evaluate TimelyDAgger on a challenging manipulation task
comprising three stages: \textbf{(1) cable pickup},
\textbf{(2) handover between grippers}, and
\textbf{(3) connector insertion into the target board}
(Fig.~\ref{fig:real_world_tasks}).
The base policy is trained with the board at a fixed position,
while OOD conditions introduce different displacements from
this position (Fig.~\ref{fig:real_world_setup}). The resulting capability gap lies primarily
in the final insertion stage, which requires precise connector
alignment at an unfamiliar target location.

We compare Offline BC, Human-Gated DAgger, and TimelyDAgger
under the same limited expert action budget at each OOD position.
Fig.~\ref{fig:real_world_success} reports results from selected
sequences of 30 consecutive trials per method.
TimelyDAgger achieves the highest success rate under both
conditions, followed by Human-Gated DAgger and Offline BC.

These differences are consistent with demonstration composition.
Offline BC allocates substantial supervision to pickup and handover,
while late takeover in Human-Gated DAgger requires the expert
to realign the connector and retry insertion after an error.
TimelyDAgger aims to focus supervision on insertion through timely
takeover. Fig.~\ref{fig:real_world_tasks} illustrates how expert
correction after a late takeover prompts FTA to lower the threshold,
enabling earlier assistance in the next rollout. The higher success
of TimelyDAgger under both OOD conditions supports the value of
takeover timing for collecting useful supervision.

\section{Conclusion}
We identify expert takeover timing as a key factor in robot-gated DAgger, shaping the quality of collected supervision and the resulting policy improvement.
TimelyDAgger combines Bridge-PCA monitoring with expert feedback
to refine takeover timing and collect more useful supervision.
Our evaluation framework
and TASR further provide tools for assessing takeover decisions and
supervision quality. We hope this work inspires advances in
timely expert supervision, data collection, and learning algorithms,
enabling robots to make better use of limited demonstrations and
acquire new capabilities more reliably and efficiently.

\bibliographystyle{IEEEtran}
\bibliography{references}

@IEEEtranBSTCTL{compactrefs,
  CTLuse_forced_etal       = "yes",
  CTLmax_names_forced_etal = "6",
  CTLnames_show_etal       = "1",
  CTLuse_url              = "no"
}

@inproceedings{pmlr-v15-ross11a,
  author    = {Ross, Stephane and Gordon, Geoffrey and Bagnell, Drew},
  title     = {A Reduction of Imitation Learning and Structured Prediction to No-Regret Online Learning},
  booktitle = {Proceedings of the Fourteenth International Conference on Artificial Intelligence and Statistics},
  year      = {2011},
  volume    = {15},
  pages     = {627--635}
}

@inproceedings{kelly2019hgdagger,
  author    = {Michael Kelly and Chelsea Sidrane and Katherine Driggs-Campbell and Mykel J. Kochenderfer},
  title     = {HG-DAgger: Interactive Imitation Learning with Human Experts},
  booktitle = {Proc. IEEE Int. Conf. Robot. Autom. (ICRA)},
  year      = {2019},
  pages     = {8077--8083},
  doi       = {10.1109/ICRA.2019.8793698}
}

@misc{zhang2016queryefficientimitationlearningendtoend,
  author = {Jiakai Zhang and Kyunghyun Cho},
  title  = {Query-Efficient Imitation Learning for End-to-End Autonomous Driving},
  year   = {2016},
  note   = {arXiv:1605.06450}
}

@misc{menda2019ensembledaggerbayesianapproachsafe,
  author = {Kunal Menda and Katherine Driggs-Campbell and Mykel J. Kochenderfer},
  title  = {{EnsembleDAgger}: A Bayesian Approach to Safe Imitation Learning},
  year   = {2019},
  note   = {arXiv:1807.08364}
}

@inproceedings{pmlr-v164-hoque22a,
  author    = {Hoque, Ryan and Balakrishna, Ashwin and Novoseller, Ellen and Wilcox, Albert and Brown, Daniel S. and Goldberg, Ken},
  title     = {{ThriftyDAgger}: Budget-Aware Novelty and Risk Gating for Interactive Imitation Learning},
  booktitle = {Proceedings of the 5th Conference on Robot Learning},
  year      = {2022},
  volume    = {164},
  pages     = {598--608}
}

@inproceedings{hoque2021lazydagger,
  author    = {Ryan Hoque and Ashwin Balakrishna and Carl Putterman and Michael Luo and Daniel S. Brown and Daniel Seita and Brijen Thananjeyan and Ellen Novoseller and Ken Goldberg},
  title     = {LazyDAgger: Reducing Context Switching in Interactive Imitation Learning},
  booktitle = {Proc. IEEE Int. Conf. Autom. Sci. Eng. (CASE)},
  year      = {2021},
  pages     = {502--509},
  doi       = {10.1109/CASE49439.2021.9551469}
}

@inproceedings{lee2024diff,
  author    = {Lee, Sung-Wook and Kang, Xuhui and Kuo, Yen-Ling},
  title     = {{Diff-DAgger}: Uncertainty Estimation with Diffusion Policy for Robotic Manipulation},
  booktitle = {International Conference on Robotics and Automation (ICRA)},
  year      = {2025},
  pages = {4845--4852},
doi   = {10.1109/ICRA55743.2025.11127730}
}

@misc{he2025uncertaintycomesfreehumanintheloop,
  author = {Zhanpeng He and Yifeng Cao and Matei Ciocarlie},
  title  = {Uncertainty Comes for Free: Human-in-the-Loop Policies with Diffusion Models},
  year   = {2025},
  note   = {arXiv:2503.01876}
}

@misc{tang2026autointervene,
  author = {Tang, Jinhe and Zhi, Weiming},
  title  = {{AutoIntervene}: Calibrated Intervention for Action-Chunking Imitation Learning Policies},
  year   = {2026},
  note   = {arXiv:2608.07065}
}

@inproceedings{NEURIPS2018_abdeb6f5,
  author    = {Lee, Kimin and Lee, Kibok and Lee, Honglak and Shin, Jinwoo},
  title     = {A Simple Unified Framework for Detecting Out-of-Distribution Samples and Adversarial Attacks},
  booktitle = {Advances in Neural Information Processing Systems},
  year      = {2018},
  volume    = {31}
}

@inproceedings{pmlr-v162-sun22d,
  author    = {Sun, Yiyou and Ming, Yifei and Zhu, Xiaojin and Li, Yixuan},
  title     = {Out-of-Distribution Detection with Deep Nearest Neighbors},
  booktitle = {Proceedings of the 39th International Conference on Machine Learning},
  year      = {2022},
  volume    = {162},
  pages     = {20827--20840}
}

@inproceedings{XuC2-RSS-25,
  author    = {Chen Xu AND Tony Khuong Nguyen AND Emma Dixon AND Christopher Rodriguez AND Patrick Miller AND Robert Lee AND Paarth Shah AND Rares Andrei Ambrus AND Haruki Nishimura AND Masha Itkina},
  title     = {{Can We Detect Failures Without Failure Data? Uncertainty-Aware Runtime Failure Detection for Imitation Learning Policies}},
  booktitle = {Proceedings of Robotics: Science and Systems},
  year      = {2025},
  doi       = {10.15607/RSS.2025.XXI.073},
  note      = {doi: 10.15607/RSS.2025.XXI.073}
}

@inproceedings{Zhou_2026_CVPR,
  author    = {Zhou, Shijie and Zhu, Bin and Yang, Jiarui and Zhao, Xiangyu and Chen, Jingjing and Jiang, Yu-Gang},
  title     = {{RC-NF}: Robot-Conditioned Normalizing Flow for Real-Time Anomaly Detection in Robotic Manipulation},
  booktitle = {Proceedings of the IEEE/CVF Conference on Computer Vision and Pattern Recognition (CVPR)},
  year      = {2026},
  pages     = {43050--43060}
}

@inproceedings{pmlr-v270-agia25a,
  author    = {Agia, Christopher and Sinha, Rohan and Yang, Jingyun and Cao, Ziang and Antonova, Rika and Pavone, Marco and Bohg, Jeannette},
  title     = {Unpacking Failure Modes of Generative Policies: Runtime Monitoring of Consistency and Progress},
  booktitle = {Proceedings of The 8th Conference on Robot Learning},
  year      = {2025},
  volume    = {270},
  pages     = {689--723}
}

@inproceedings{NEURIPS2025_0b7cb3b8,
  author    = {R\"{o}mer, Ralf and Kobras, Adrian and Worbis, Luca and Schoellig, Angela},
  title     = {Failure Prediction at Runtime for Generative Robot Policies},
  booktitle = {Advances in Neural Information Processing Systems},
  year      = {2025},
  volume    = {38, Main Conference},
  pages     = {8509--8548},
  doi       = {10.52202/085713-0260},
  note      = {doi: 10.52202/085713-0260}
}

@misc{zheng2026rewindilonlinefailuredetection,
  author = {Gehan Zheng and Sanjay Seenivasan and Matthew Johnson-Roberson and Weiming Zhi},
  title  = {{Rewind-IL}: Online Failure Detection and State Respawning for Imitation Learning},
  year   = {2026},
  note   = {arXiv:2604.16683}
}

@inproceedings{ICLR2025_70a06501,
  author    = {Duan, Jiafei and Pumacay, Wilbert and Kumar, Nishanth and Wang, Yi Ru and Tian, Shulin and Yuan, Wentao and Krishna, Ranjay and Fox, Dieter and Mandlekar, Ajay and Guo, Yijie},
  title     = {{AHA}: A Vision-Language-Model for Detecting and Reasoning Over Failures in Robotic Manipulation},
  booktitle = {International Conference on Learning Representations},
  year      = {2025},
  volume    = {2025},
  pages     = {45493--45517}
}

@inproceedings{lin2026failsafe,
  author    = {Zijun Lin and Jiafei Duan and Haoquan Fang and Dieter Fox and Ranjay Krishna and Cheston Tan and Bihan Wen},
  title     = {FailSafe: Reasoning and Recovery from Failures in Vision-Language-Action Models},
  booktitle = {Proc. IEEE/RSJ Int. Conf. Intell. Robots Syst. (IROS)},
  year      = {2026}
}

@misc{ma2026cyclevla,
  author = {Ma, Chenyang and Lu, Kai and Yang, Guangyu and Liu, Jiuming and Xu, Shitong and Byrne, Bill and Havoutis, Ioannis and Trigoni, Niki and Markham, Andrew},
  title  = {{CycleVLA}: Proactive Self-Correcting Vision-Language-Action Models via Subtask Backtracking and Minimum Bayes Risk Decoding},
  year   = {2026},
  note   = {arXiv:2601.02295}
}

@inproceedings{rolland2026failure,
  author    = {Rolland, Quentin and Mayran de Chamisso, Fabrice and Mouret, Jean-Baptiste},
  title     = {Failure Identification in Imitation Learning via Statistical and Semantic Filtering},
  booktitle = {IEEE International Conference on Robotics and Automation (ICRA)},
  year      = {2026}
}

@inproceedings{NEURIPS2025_392d0d05,
  author    = {Gu, Qiao and Ju, Yuanliang and Sun, Shengxiang and Gilitschenski, Igor and Nishimura, Haruki and Itkina, Masha and Shkurti, Florian},
  title     = {{SAFE}: Multitask Failure Detection for Vision-Language-Action Models},
  booktitle = {Advances in Neural Information Processing Systems},
  year      = {2025},
  volume    = {38, Main Conference},
  pages     = {40041--40076},
  doi       = {10.52202/085713-1337},
}

@misc{park2026hideandseektrajectoriesdiscoveringfailure,
  author = {Seongheon Park and Wendi Li and Changdae Oh and Samuel Yeh and Zsolt Kira and Michael Hagenow and Sharon Li},
  title  = {{Hide-and-Seek} in Trajectories: Discovering Failure Signals for {VLA} Runtime Monitoring},
  year   = {2026},
  note   = {arXiv:2605.30834}
}

@misc{mandlekar2020humanintheloopimitationlearningusing,
  author = {Ajay Mandlekar and Danfei Xu and Roberto Mart{\'i}n-Mart{\'i}n and Yuke Zhu and Li Fei-Fei and Silvio Savarese},
  title  = {Human-in-the-Loop Imitation Learning using Remote Teleoperation},
  year   = {2020},
  note   = {arXiv:2012.06733}
}

@inproceedings{taomaniskill3,
    author  = {Stone Tao and Fanbo Xiang and Arth Shukla and Yuzhe Qin and Xander Hinrichsen and Xiaodi Yuan and Chen Bao and Xinsong Lin and Yulin Liu and Tse-kai Chan and Yuan Gao and Xuanlin Li and Tongzhou Mu and Nan Xiao and Arnav Gurha and Viswesh Nagaswamy Rajesh and Yong Woo Choi and Yen-Ru Chen and Zhiao Huang and Roberto Calandra and Rui Chen and Shan Luo and Hao Su},
  title = {Demonstrating {GPU} Parallelized Robot Simulation
           and Rendering for Generalizable Embodied {AI}
           with {ManiSkill3}},
  booktitle = {Proceedings of Robotics: Science and Systems},
  year = {2025},
  doi = {10.15607/RSS.2025.XXI.021}
}

@misc{intelligence2025pi05visionlanguageactionmodelopenworld,
  author = {{Physical Intelligence} and Kevin Black and Noah Brown and James Darpinian and Karan Dhabalia and Danny Driess and Adnan Esmail and Michael Equi and Chelsea Finn and Niccolo Fusai and Manuel Y. Galliker and Dibya Ghosh and Lachy Groom and Karol Hausman and Brian Ichter and Szymon Jakubczak and Tim Jones and Liyiming Ke and Devin LeBlanc and Sergey Levine and Adrian Li-Bell and Mohith Mothukuri and Suraj Nair and Karl Pertsch and Allen Z. Ren and Lucy Xiaoyang Shi and Laura Smith and Jost Tobias Springenberg and Kyle Stachowicz and James Tanner and Quan Vuong and Homer Walke and Anna Walling and Haohuan Wang and Lili Yu and Ury Zhilinsky},
  title  = {$\pi_{0.5}$: a Vision-Language-Action Model with Open-World Generalization},
  year   = {2025},
  note   = {arXiv:2504.16054}
}

@misc{kim24openvla,
      author={Moo Jin Kim and Karl Pertsch and Siddharth Karamcheti and Ted Xiao and Ashwin Balakrishna and Suraj Nair and Rafael Rafailov and Ethan Foster and Grace Lam and Pannag Sanketi and Quan Vuong and Thomas Kollar and Benjamin Burchfiel and Russ Tedrake and Dorsa Sadigh and Sergey Levine and Percy Liang and Chelsea Finn},
  title  = {{OpenVLA}: An Open-Source Vision-Language-Action Model},
  year   = {2024},
  note   = {arXiv:2406.09246}
}

@misc{zheng2025x,
  author = {Zheng, Jinliang and Li, Jianxiong and Wang, Zhihao and Liu, Dongxiu and Kang, Xirui and Feng, Yuchun and Zheng, Yinan and Zou, Jiayin and Chen, Yilun and Zeng, Jia and others},
  title  = {{X-VLA}: Soft-Prompted Transformer as Scalable Cross-Embodiment Vision-Language-Action Model},
  year   = {2025},
  note   = {arXiv:2510.10274}
}

@misc{firouzkouhi2025sampleefficientexpertquerycontrol,
  author = {Arad Firouzkouhi and Omid Mirzaeedodangeh and Lars Lindemann},
  title  = {Sample-Efficient Expert Query Control in Active Imitation Learning via Conformal Prediction},
  year   = {2025},
  note   = {arXiv:2512.00453}
}

@misc{seligmann2026vlafailefficienttaskfailure,
  author = {Florian Seligmann and Emiliyan Gospodinov and Enes Ulas Dincer and Gerhard Neumann},
  title  = {{VLA-FAIL}: Efficient Task Failure Detection for Finetuned Vision-Language-Action Models},
  year   = {2026},
  note   = {arXiv:2606.21386}
}

@inproceedings{Davis_2006,
  author    = {Davis, Jesse and Goadrich, Mark},
  title     = {The relationship between Precision-Recall and {ROC} curves},
  booktitle = {Proceedings of the 23rd international conference on Machine learning  - ICML '06},
  year      = {2006},
  pages     = {233--240},
  doi       = {10.1145/1143844.1143874},
}

@article{Chernova_2009,
  author  = {Sonia Chernova and Manuela Veloso},
  title   = {Interactive Policy Learning through Confidence-Based Autonomy},
  journal = {Journal of Artificial Intelligence Research},
  year    = {2009},
  volume  = {34},
  pages   = {1--25},
  doi     = {10.1613/jair.2584},
  note    = {doi: 10.1613/jair.2584}
}

@article{JMLR:v15:judah14a,
  author  = {Kshitij Judah and Alan P. Fern and Thomas G. Dietterich and Prasad Tadepalli},
  title   = {Active Imitation Learning: Formal and Practical Reductions to {I.I.D.} Learning},
  journal = {Journal of Machine Learning Research},
  year    = {2014},
  volume  = {15},
  number  = {120},
  pages   = {4105--4143}
}

\end{document}